\documentclass[11pt]{article}

\usepackage[preprint]{acl}

\usepackage{times}
\usepackage{latexsym}

\usepackage[T1]{fontenc}

\usepackage[utf8]{inputenc}

\usepackage{microtype}

\usepackage{inconsolata}

\usepackage{graphicx}
\usepackage{amssymb}
\usepackage{subcaption}
\usepackage{amsmath}
\usepackage{booktabs}
\usepackage{longtable}
\usepackage{colortbl}
\usepackage{multirow}
\usepackage{arydshln}
\usepackage[skins,breakable,listings]{tcolorbox}
\definecolor{groupblue}{HTML}{CDE8E5}
\definecolor{checkgreen}{HTML}{2E7D32}
\definecolor{crossred}{HTML}{B23A48}
\definecolor{promptgray}{HTML}{666666}
\newtcblisting{promptbox}[1]{
  enhanced,
  breakable,
  listing only,
  width=\linewidth,
  arc=2pt,
  boxrule=0.7pt,
  colframe=promptgray,
  colback=white,
  colbacktitle=promptgray,
  coltitle=white,
  fonttitle=\bfseries\small,
  title={#1},
  left=1mm,
  right=1mm,
  top=1mm,
  bottom=1mm,
  listing options={
    basicstyle=\ttfamily\tiny,
    columns=fullflexible,
    keepspaces=true,
    breaklines=true,
    breakatwhitespace=true,
    breakautoindent=false,
    breakindent=0pt,
    showstringspaces=false
  }
}
\newcommand{\method}{FaithMed}
\newcommand{\cmark}{\textcolor{checkgreen}{\checkmark}}
\newcommand{\xmark}{\textcolor{crossred}{\(\times\)}}
\title{FaithMed: Training LLMs For Faithful Evidence-Based Medical Reasoning}
\author{
  \textbf{Zhiyun Zhang\textsuperscript{1$\ast$}},
  \textbf{Liwen Sun\textsuperscript{1$\ast$}},
  \textbf{Xiang Qian\textsuperscript{2}},
  \textbf{Chenyan Xiong\textsuperscript{1,3}}
\\
  \textsuperscript{1}Carnegie Mellon University \quad
  \textsuperscript{2}Stanford University School of Medicine \quad
  \textsuperscript{3}Xlue
\\
  $\ast$Equal contribution \\
  \small{\{zhiyunz, liwens, cx\}@andrew.cmu.edu, xqian@stanford.edu}
}

\begin{document}
\maketitle

\begin{abstract}
\label{sec:abstract}
Faithful reasoning is essential in medicine, where clinical decisions require
transparent justification grounded in reliable evidence. Current medical LLMs
either lack active access to evidence or use retrieved evidence without
supervising how it should be appraised and applied during reasoning.
To address this, we formalize evidence-based medicine principles as
process-level criteria and introduce \method, a framework that combines
clinician-designed, automatically refined rubrics with reinforcement learning
using step-level process reward assignment and advantage grouping.
Across seven medical benchmarks, \method{} improves over agentic-search
baselines (+9\% on average) and outcome-only RL (+5.8\%), while raising average
evidence-based medicine rubric scores over agentic-search Qwen3 baselines
(+15.5\%).
This work demonstrates that explicit step-level supervision can improve both
task success and the faithfulness of the reasoning process.
Code is available at \url{https://github.com/cxcscmu/FaithMed}.
\end{abstract}

\section{Introduction } 
\label{sec:intro}

In medicine, reliable clinical decision-making depends on reasoning that connects patient-specific information with trustworthy evidence and makes its justification transparent~\citep{schwartz2006med}. Evidence-based medical reasoning operationalizes this goal, requiring physicians to synthesize patient history, clinical findings, and medical guidelines to support accurate diagnoses and treatment planning~\citep{sackett1996evidence}. In real-world healthcare settings, this reasoning process is often complex, time-consuming, and vulnerable to incomplete documentation, noisy observations, and inconsistent evidence across longitudinal patient records~\citep{gruppen2017clinical,temsah2023chatgpt}.

Recent advances in large language models (LLMs) have demonstrated strong capabilities in medical question answering (QA) and clinical reasoning, motivating the development of automated systems that can assist clinicians in interpreting medical evidence and supporting healthcare decisions~\citep{jin-etal-2019-pubmedqa,singhal2023expertlevelmedicalquestionanswering,sun2024edcopilotreduceemergencydepartment}. However, ensuring that these systems produce faithful and evidence-grounded reasoning remains critical for improving reliability, transparency, and patient safety in high-stakes clinical environments~\citep{singhal2023large,saab2024capabilitiesgeminimodelsmedicine}.

\begin{figure}[t]
\centering
\includegraphics[width=\columnwidth]{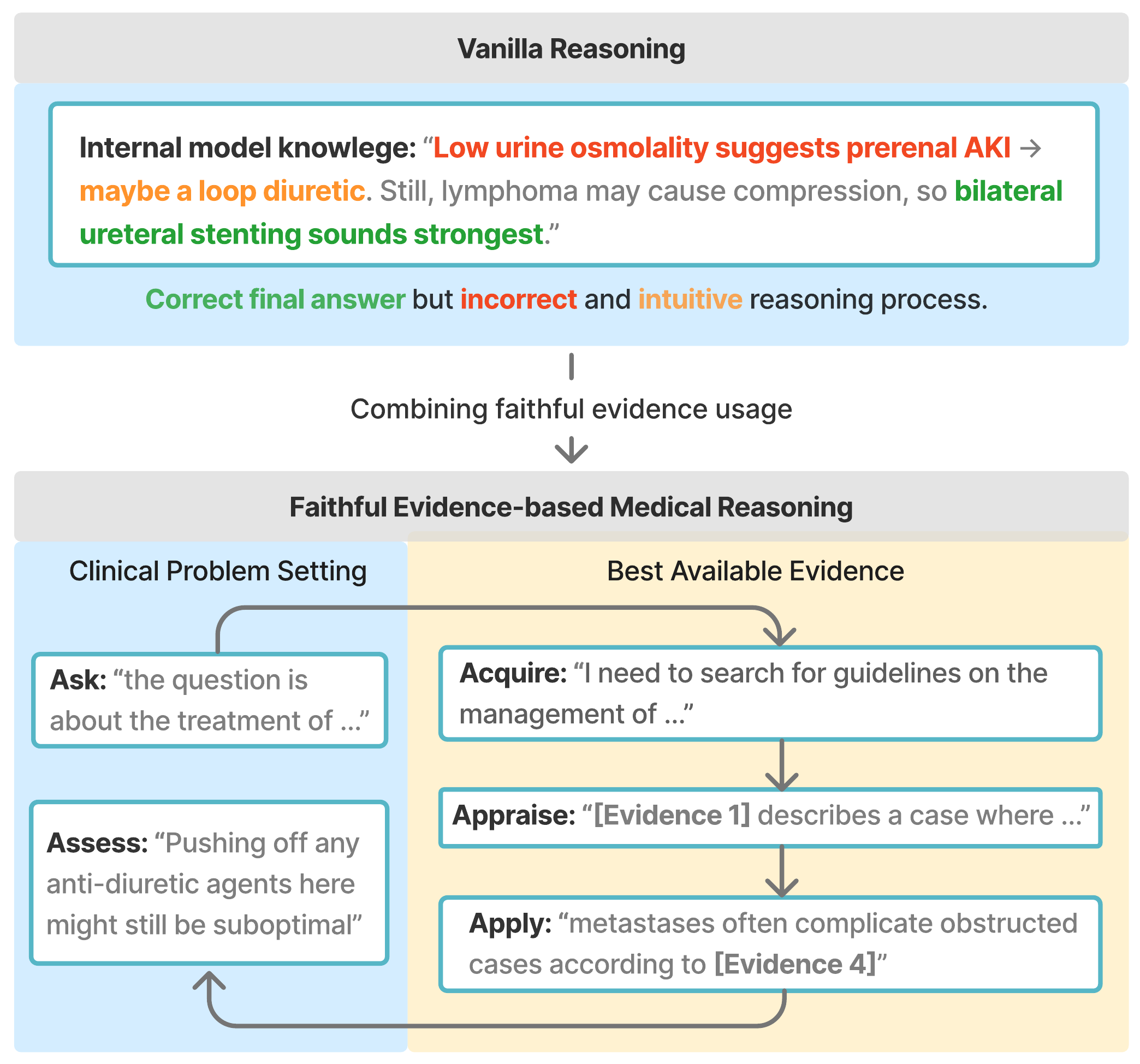}
\caption{Comparing unfaithful reasoning with evidence-based medical reasoning.}
\label{fig:motivation}
\end{figure}

To improve factual grounding and attribution, prior work has explored retrieval-augmented generation (RAG), which combines LLMs with external trustworthy medical knowledge sources to generate verifiable responses with supporting citations~\citep{wang-etal-2025-medcite,xiong2024benchmarkingretrievalaugmentedgenerationmedicine,gao-etal-2023-enabling}. While these approaches can improve answer correctness and attribution, the reasoning process itself may still be unfaithful to the supporting evidence. In particular, models may generate plausible rationales that drift from the patient context, cite irrelevant evidence, or make claims that are not fully supported by the retrieved information, as illustrated in Figure~\ref{fig:motivation}. Although the final answer is correct, the underlying reasoning process can still be flawed and unreliable.

To address this limitation, we introduce \method, a framework for faithful evidence-based medical reasoning that trains LLMs to generate evidence-supported reasoning through rubric-guided reinforcement learning. Inspired by real-world evidence-based medicine workflows---\textit{ask}, \textit{acquire}, \textit{appraise}, \textit{apply}, and \textit{assess}---we collaborate with clinicians to design initial rubrics that evaluate reasoning quality across these five dimensions, then automatically refine them to reduce redundancy and improve discriminativeness. We then integrate these refined rubrics into a step-level reinforcement learning framework with rubric-guided rewards. Instead of producing unconstrained rationales, \method~encourages the model to follow clinically meaningful reasoning by grounding intermediate reasoning steps in supporting evidence and generating accurate answers consistent with the evidence-based reasoning process.

Across seven medical benchmarks, \method{} improves answer accuracy by 9\% over agentic-search methods and by 10.8\% over outcome-only RL baselines, while achieving a 15.5\% gain in evidence-based medicine rubric scores. These gains suggest that faithful process supervision improves not only the reasoning trace but also final task performance. We further investigate different faithful reasoning injection strategies by comparing episode-level and step-level process rewards in reinforcement learning, and show that fine-grained step-level rewards derived from individual rubrics achieve superior performance. Moreover, our analysis reveals that episode-level process rewards can incorrectly assign credit across step groups, leading to suboptimal optimization. Finally, clinician-annotated case studies demonstrate that the reasoning produced by our method aligns more closely with the corresponding evidence-based medicine dimensions than strong baselines with agentic search.

Our main contributions can be summarized
as follows:
\begin{itemize}

\item We propose a framework for faithful evidence-based medical reasoning that enables LLMs to generate clinically grounded responses with evidence-supported reasoning processes.

\item We develop clinician-designed and automatically refined rubrics for faithful reasoning evaluation, and use them for reinforcement learning with step-level process reward assignment and advantage grouping.

\item We demonstrate that our method improves not only answer accuracy but also reasoning faithfulness, outperforming existing baselines across multiple medical QA benchmarks.
\end{itemize}

\section{Related Work}
\label{sec:related}

\begin{figure*}[t]
\centering
\includegraphics[width=\textwidth]{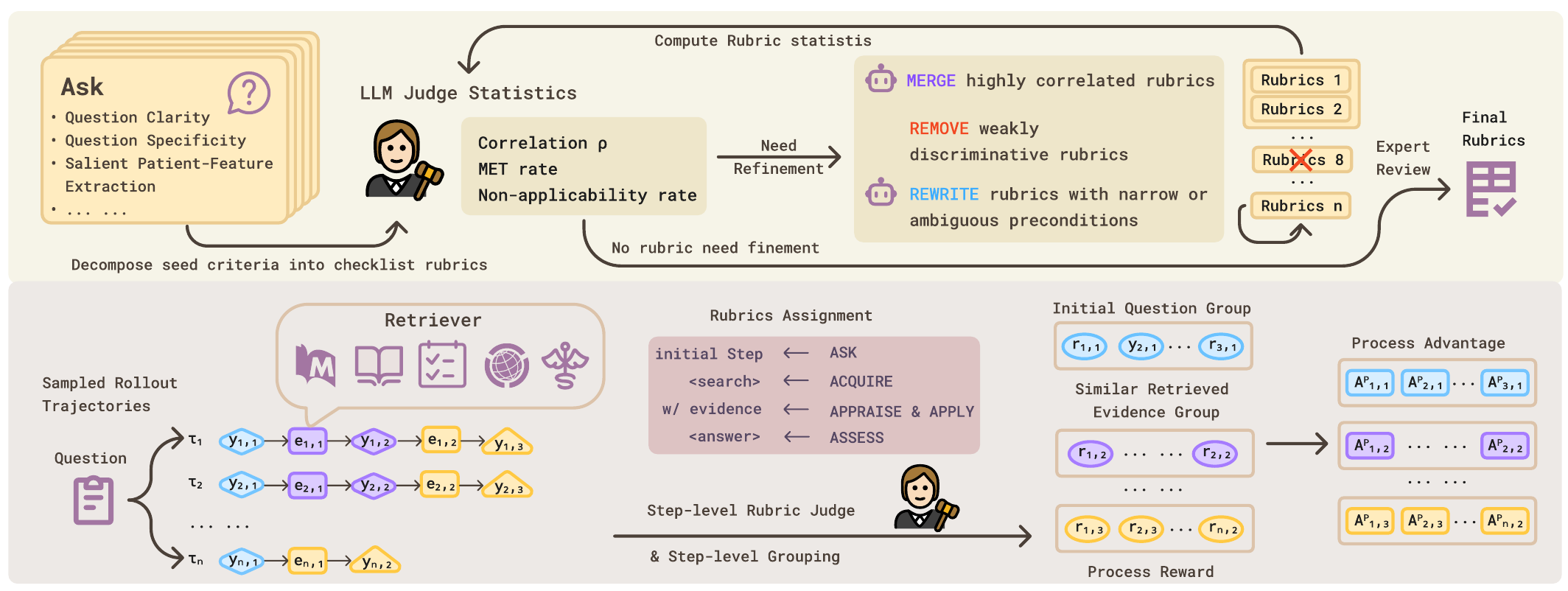}
\caption{Method overview. We first design and refine rubrics for evaluating the reasoning process across five evidence-based medicine dimensions. We then use these rubrics to construct step-level process rewards, which are integrated with outcome rewards for policy optimization.}
\label{fig:method-overview}
\end{figure*}

\paragraph{Medical RAG.}
Large language models have demonstrated promising performance on medical question answering benchmarks, as shown by systems such as Med-PaLM \citep{singhal2023large}, Meditron \citep{chen2023meditron70bscalingmedicalpretraining}, and Med-Gemini \citep{saab2024capabilitiesgeminimodelsmedicine}. However, these models remain prone to hallucination, motivating the adoption of retrieval-augmented generation (RAG) to ground responses in retrieved medical documents and improve factuality \citep{lewis2021retrievalaugmentedgenerationknowledgeintensivenlp,xiong2024benchmarkingretrievalaugmentedgenerationmedicine,info15080494,sun2025factawaremultimodalretrievalaugmentation}. Recent work such as MedCite~\citep{wang-etal-2025-medcite} further emphasizes response verifiability by adapting citation generation and retrieval strategies~\citep{gao-etal-2023-enabling,Huo_2023,ye2024effectivelargelanguagemodel} to the medical domain with specialized evaluation protocols, improving attribution to external medical knowledge sources. Nevertheless, existing approaches primarily focus on the factuality of final responses, while largely overlooking the faithfulness of the intermediate reasoning process, which may still lead to unreliable clinical conclusions.

\paragraph{Medical Reasoning.}
Clinical decision-making often requires complex multi-step reasoning over patient evidence during diagnosis and treatment planning. Prior work such as MedPrompt~\citep{nori2023generalistfoundationmodelsoutcompete} leverages Chain-of-Thought (CoT) prompting \citep{wei2023chainofthoughtpromptingelicitsreasoning} to improve medical reasoning capabilities of general-purpose LLMs during inference. Similarly, Huatuo-o1 \citep{chen2024huatuogpto1medicalcomplexreasoning} iteratively improves multi-step reasoning trajectories through verifier-guided feedback. 

More recently, agentic medical systems have begun integrating reinforcement learning with RAG and process reward modeling using LLM-as-a-judge frameworks to improve reasoning quality~\citep{jin2025searchr1,yun2025medprmmedicalreasoningmodels,jiang2025meds3medicalslowthinking}. Different from prior work, our method incorporates realistic evidence-based medicine workflows into a reinforcement learning framework with fine-grained step-level rewards. Guided by clinician-curated rubrics, our framework explicitly promotes faithful and trustworthy reasoning processes that remain grounded in supporting medical evidence throughout the path toward the final clinical conclusion.

\section{Preliminaries}
\label{sec:preliminaries}

\paragraph{Reasoning Environment.}
We formulate evidence-based medical reasoning as an agent--environment interaction.
Given a clinical question $x$, the initial state $s_1$ is conditioned on $x$. At
each turn $k$, the model observes $s_k$ and generates $y_k=(t_k, a_k)$, where $t_k$
denotes the reasoning trace and $a_k \in \{\texttt{search}, \texttt{answer}\}$ is
the action~\citep{jin2025searchr1}. A \texttt{search} action retrieves external
evidence $e_k = \{e_{k,j}\}_{j}$, transitioning to the next state $s_{k+1}$ and
yielding reward $r_k$. The episode terminates when the model selects the
\texttt{answer} action, producing a $T$-turn trajectory
$\tau = \{(s_k, t_k, a_k, r_k)\}_{k=1}^{T}$. To enable citation grounding, each
retrieved document $e_{k,j}$ is serialized within an \texttt{<information>} block
and assigned the identifier \texttt{[$Tk$-$Ej$]}, which the model is prompted to
reference in subsequent reasoning.

\paragraph{Outcome-Reward RL.}
The outcome reward for a trajectory $\tau$ is defined as
$R^O(\tau) = \sum_{k=1}^{T} r^O_k$. In the standard GRPO
setting, reward is assigned only at the end of the episode~\citep{shao2024deepseekmath}:
$r^O_k = 0$ for $k < T$, while $r^O_T$ is a binary signal indicating the
correctness of the final answer. For each question, GRPO samples $N$ candidate
episodes $\{\tau_i\}_{i=1}^{N}$ from the policy $\pi_\theta(y_k \mid s_k, x)$
and computes group-normalized episode rewards:
\begin{equation}
\label{eq:outcome-advantage}
A_i^O =
\frac{R^O(\tau_i) - \mathrm{mean}\bigl(\{R^O(\tau_j)\}_{j=1}^{N}\bigr)}
{\mathrm{std}\bigl(\{R^O(\tau_j)\}_{j=1}^{N}\bigr)}.
\end{equation}
This yields an episode-level outcome advantage shared across the entire trajectory.

\section{\method: Towards Fine-Grained Process Reward in Medical Reasoning}
\label{sec:Methods}

We present \method, which trains LLMs with rubric-guided step-level process rewards for
evidence-based medical reasoning. We first describe how the evaluation criteria
are designed and refined in Section~\ref{subsec:rubrics_design}. We then define
the step-level process reward in Section~\ref{subsec:process_reward} and show
how it is incorporated into policy optimization in
Section~\ref{subsec:policy_optimization}.

\subsection{Rubrics Design and Refinement}
\label{subsec:rubrics_design}

We first define the process-reward dimensions following the principles of evidence-based
medicine~\citep{sackett1996evidence,dusin2023evidence}: \textit{Ask}, \textit{Acquire},
\textit{Appraise}, \textit{Apply}, and \textit{Assess}. Specifically, (1) \textit{Ask}
evaluates whether the information need is formulated as a clear clinical question;
(2) \textit{Acquire} measures whether the model retrieves relevant external evidence;
(3) \textit{Appraise} evaluates whether the retrieved evidence is critically assessed for
relevance and support; (4) \textit{Apply} measures whether the evidence is integrated with the
clinical context; and (5) \textit{Assess} evaluates whether the overall reasoning process and
final decision are reviewed. For each dimension, we first design seed criteria and
then decompose them into binary checklist-style rubrics following
CheckEval~\citep{lee-etal-2025-checkeval}. 

To reduce redundancy and eliminate uninformative criteria, we use teacher-generated medical QA reasoning trajectories to iteratively refine the curated rubric set until convergence. Specifically, we adopt a metric-driven recursive rubric-refinement procedure. In each iteration, an LLM judge scores all trajectory--rubric pairs, after which we compute rubric-level statistics, including non-applicability rate, \textsc{Met} rate, and pairwise Pearson correlation between rubrics.

These rubric-level statistics are then used to determine the refinement action:
(1) \textsc{Merge} rubrics with high pairwise correlation, as they capture
overlapping reasoning behaviors; (2) \textsc{Remove} weakly discriminative
rubrics with nearly constant labels, including rubrics that are almost always
\textsc{Met} or \textsc{Unmet}; and (3) \textsc{Rewrite} rubrics
with high non-applicability rates, which typically reflect overly narrow or
ambiguous preconditions. The refinement process terminates when no active rubric satisfies
any refinement criterion. \textsc{Merge} and \textsc{Rewrite} operations are automatically
performed by an LLM editor. Finally, a medical expert reviews the
refined rubric set and performs final edits, yielding the rubrics used
for process evaluation; the final rubric set is provided in Appendix
\ref{app:rubric-details}.

\subsection{Step-level Process Reward }
\label{subsec:process_reward}

Outcome rewards only indicate whether the final answer is correct, but provide
limited guidance on whether intermediate reasoning decisions are faithful to the
available evidence. We therefore construct a process reward for
each generated step $y_k=(t_k,a_k)$ by judging the step against the rubric
dimensions that are relevant to its local context. Let $e_0=\emptyset$. For a
step $k$, we assign a dimension set $D_k$ as follows: the initial step includes
\textit{Ask}; a \texttt{search} action includes \textit{Acquire}; if the model
has observed non-empty retrieved information $e_{k-1}$, the step includes
\textit{Appraise} and \textit{Apply}; and an \texttt{answer} action includes
\textit{Assess}. These assignments are additive, so a step can be evaluated by
multiple dimensions.

Then, we construct the applicable rubric set and evaluate the corresponding reasoning step:
\begin{align}
\mathcal{R}_k&=\bigcup_{d\in D_k}\mathcal{R}_d,\\
J_k&=J(y_k,e_{k-1},\mathcal{R}_k)\in\{0,1\}^{|\mathcal{R}_k|},
\end{align}
where $\mathcal{R}_d$ denotes the refined rubrics for dimension $d$, and
$J_k$ denotes the binary rubric scores returned by the judge for step $k$.
For trajectory $i$, we aggregate the applicable rubric scores into a scalar
step-level process reward $r^P_{i,k}=\mathrm{Agg}(J_{i,k})$, where
$\mathrm{Agg}(\cdot)$ denotes mean-centered rubric aggregation.

To obtain comparable process rewards, we group reasoning steps with similar
observation contexts and retrieved evidence as anchor-state, following the strategy of GiGPO~\citep{feng2025gigpo}. The process advantage is computed using group-normalized step-level rewards:
\begin{equation}
\label{eq:process-advantage}
A^{P}_{i,k}=
\frac{
r^{P}_{i,k}-\mathrm{mean}\!\left(\mathcal{G}(\tilde{s}_{i,k})\right)
}{
\mathrm{std}\!\left(\mathcal{G}(\tilde{s}_{i,k})\right)+\epsilon
},
\end{equation}
where $\mathcal{G}(\tilde{s}_{i,k})$ denotes the multiset of process rewards in
the anchor group associated with anchor state $\tilde{s}_{i,k}$. This yields fine-grained comparative process feedback, where a search or answer
step is rewarded when it better satisfies the relevant rubrics than alternative
steps under a similar observation context.

\subsection{FaithMed Policy Optimization} 
\label{subsec:policy_optimization}

Outcome-only RL assigns the same episode-level signal to every step in a
trajectory. In \method, the final-answer reward $r^O$ and rubric-guided
process reward $r^P$ are treated as separate supervision signals:
$r^O$ defines the episode-level outcome advantage $A_i^O$ in
Eq.~\ref{eq:outcome-advantage}, while $r^P$ defines the step-level process
advantage $A^P_{i,k}$ in Eq.~\ref{eq:process-advantage}. The advantage for step
$k$ in trajectory $i$ is computed as
\begin{equation}
A_{i,k}=A^{O}_i+\lambda A^{P}_{i,k},
\end{equation}
where $\lambda$ controls the contribution of rubric-guided process feedback.
The first term encourages trajectories that reach the correct final answer,
while the second term provides additional credit to intermediate reasoning steps
that better satisfy the process rubrics under a similar evidence context. We then optimize $A_{i,k}$ using the
standard GRPO clipped objective over the generated tokens in step $y_{i,k}$,
together with the standard reference-policy KL penalty. This
enables \method\ to jointly optimize final-answer correctness and intermediate
evidence-based medicine behaviors.

\begin{table*}[t]
\centering
\small
\setlength{\tabcolsep}{3.5pt}
\renewcommand{\arraystretch}{1.12}
\caption{Main benchmark results. \method{} here uses the no-SFT recipe. GRPO uses outcome rewards only. $^\S$ denotes reference LLM values reported by Medmarks-Verifiable~\citep{warner2026medmarks}. $^\dagger$ denotes in-domain datasets and $^\ast$ denotes out-of-domain datasets. }
\label{tab:main-results}
\resizebox{\textwidth}{!}{%
\begin{tabular}{llcccccccc}
\toprule
\multicolumn{2}{l}{\textbf{Method}} &
\textbf{MedQA$^\dagger$} &
\textbf{MedMCQA$^\dagger$} &
\textbf{HeadQA$^\dagger$} &
\begin{tabular}[c]{@{}c@{}}\textbf{MedCalc-}\\\textbf{Bench$^\dagger$}\end{tabular} &
\textbf{MedBullets$^\ast$} &
\begin{tabular}[c]{@{}c@{}}\textbf{MMLU-Pro-}\\\textbf{Health$^\ast$}\end{tabular} &
\begin{tabular}[c]{@{}c@{}}\textbf{MedXpert}\\\textbf{QA$^\ast$}\end{tabular} &
\begin{tabular}[c]{@{}c@{}}\textbf{Overall}\\\textbf{Acc.}\end{tabular} \\
\midrule
\rowcolor{groupblue}
\multicolumn{10}{l}{\textbf{Reference LLMs}} \\
\multicolumn{2}{l}{Llama 3.1 8B Instruct} & 61.8 & 57.3 & 67.6 & 21.4 & 52.2 & 56.0 & 11.8 & 46.9 \\
\multicolumn{2}{l}{Gemma 3 12B$^\S$} & 69.0 & 59.0 & 73.0 & 27.0 & 51.0 & 59.0 & 16.4 & 50.6 \\
\multicolumn{2}{l}{Qwen3 8B} & 79.4 & 66.3 & 83.4 & 44.9 & 61.8 & 69.7 & 19.2 & 60.7 \\
\multicolumn{2}{l}{MedGemma 27B$^\S$} & 83.0 & 73.0 & 84.0 & 49.0 & 75.0 & 72.0 & 19.0 & 65.0 \\
\multicolumn{2}{l}{Claude Sonnet 4.5$^\S$} & 93.0 & 80.0 & 91.0 & 76.0 & 80.0 & 84.0 & 39.4 & 77.6 \\
\multicolumn{2}{l}{DeepSeek-R1} & 94.2 & 72.8 & 93.4 & 72.7 & 73.2 & 74.2 & 42.9 & 74.8 \\
\midrule
\rowcolor{groupblue}
\multicolumn{10}{l}{\textbf{\method{}}} \\
\multirow{5}{*}{\begin{tabular}[c]{@{}l@{}}Qwen3\\1.7B\end{tabular}} & \hspace{2pt}Base & 48.9 & 48.0 & 66.2 & 16.9 & 35.7 & 43.4 & 9.3 & 38.3 \\
 & \hspace{2pt}MedRAG & \underline{51.4} & \underline{51.3} & 67.5 & 8.4 & \textbf{40.8} & \underline{48.5} & \underline{14.2} & 40.3 \\
 & \hspace{2pt}Agentic Search & 49.2 & 46.8 & 65.5 & 15.3 & 36.0 & 46.4 & 11.1 & 38.6 \\
 & \hspace{2pt}Agentic Search + GRPO & 50.9 & 50.1 & \underline{68.5} & \underline{27.1} & 31.0 & 48.1 & 11.5 & \underline{41.0} \\
 & \hspace{2pt}\textbf{\method{}}  & \textbf{61.4} & \textbf{56.9} & \textbf{75.6} & \textbf{33.2} & \underline{38.5} & \textbf{52.0} & \textbf{17.1} & \textbf{47.8} \\
\cmidrule(lr){1-10}
\multirow{5}{*}{\begin{tabular}[c]{@{}l@{}}Qwen3\\4B\end{tabular}} & \hspace{2pt}Base & 69.3 & 59.7 & 81.3 & 13.7 & 46.3 & 63.0 & 16.0 & 49.9 \\
 & \hspace{2pt}MedRAG & 70.2 & 59.8 & 83.9 & 10.3 & \underline{52.1} & 55.5 & 19.0 & 50.1 \\
 & \hspace{2pt}Agentic Search & 70.6 & \underline{63.4} & 81.5 & 29.6 & 47.3 & 55.4 & 19.6 & 52.5 \\
 & \hspace{2pt}Agentic Search + GRPO & \underline{71.3} & 63.2 & \underline{84.4} & \underline{39.0} & 49.8 & \textbf{66.1} & \underline{22.0} & \underline{56.5} \\
 & \hspace{2pt}\textbf{\method{}} & \textbf{74.8} & \textbf{64.1} & \textbf{85.7} & \textbf{56.5} & \textbf{55.6} & \underline{63.5} & \textbf{28.2} & \textbf{61.2} \\
\bottomrule
\end{tabular}%
}
\end{table*}


\begin{table}[t]
\centering
\footnotesize
\setlength{\tabcolsep}{3pt}
\renewcommand{\arraystretch}{1.12}
\caption{Process scores across five dimensions. + Ep. PR uses episode-level process reward, while \method{} uses step-level process reward.}
\label{tab:rubric-results}
\resizebox{\columnwidth}{!}{%
\begin{tabular}{@{}llrrrrr@{}}
\toprule
\textbf{Backbone} &
\textbf{Method} &
\textbf{Ask} &
\textbf{Acquire} &
\textbf{Appraise} &
\textbf{Apply} &
\textbf{Assess} \\
\midrule
\multicolumn{2}{@{}l}{DeepSeek-R1} & 94.7 & 52.7 & 48.5 & 46.2 & 14.6 \\
\midrule
\multirow{3}{*}{Qwen3 1.7B}
& Base & 82.3 & 27.1 & 23.5 & 20.6 & 6.6 \\
& + Ep. PR & 84.6 & 38.6 & 34.3 & 31.6 & 10.9 \\
& \textbf{\method{}} & \textbf{85.8} & \textbf{50.1} & \textbf{44.9} & \textbf{38.8} & \textbf{17.5} \\
\midrule
\multirow{3}{*}{Qwen3 4B}
& Base & 87.0 & 30.2 & 26.1 & 23.3 & 9.0 \\
& + Ep. PR & 86.7 & 41.4 & 36.4 & 34.0 & 12.5 \\
& \textbf{\method{}} & \textbf{90.4} & \textbf{53.9} & \textbf{48.5} & \textbf{41.4} & \textbf{19.1} \\
\bottomrule
\end{tabular}
}
\end{table}

\begin{table*}[t]
\centering
\small
\setlength{\tabcolsep}{4pt}
\renewcommand{\arraystretch}{1.12}
\caption{Training-recipe ablation. OR denotes outcome reward, PR denotes process reward, and Ep. and Step indicate episode-level and step-level PR, respectively.}
\label{tab:training-ablation}
\resizebox{\textwidth}{!}{%
\begin{tabular}{lccclcccccccc}
\toprule
\multirow{2}{*}{\textbf{Backbone}} &
\multirow{2}{*}{\textbf{SFT}} &
\multicolumn{2}{c}{\textbf{RL Rewards}} &
\multirow{2}{*}{\textbf{MedQA$^\dagger$}} &
\multirow{2}{*}{\textbf{MedMCQA$^\dagger$}} &
\multirow{2}{*}{\textbf{HeadQA$^\dagger$}} &
\multirow{2}{*}{\textbf{MedCalc-Bench$^\dagger$}} &
\multirow{2}{*}{\textbf{MedBullets$^\ast$}} &
\multirow{2}{*}{\begin{tabular}[c]{@{}c@{}}\textbf{MMLU-Pro-}\\\textbf{Health$^\ast$}\end{tabular}} &
\multirow{2}{*}{\textbf{MedXpertQA$^\ast$}} &
\multirow{2}{*}{\begin{tabular}[c]{@{}c@{}}\textbf{Overall}\\\textbf{Acc.}\end{tabular}} \\
\cmidrule(lr){3-4}
 & &
\textbf{OR} &
\textbf{PR} &
 & & & & & & & \\
\midrule
\multirow{8}{*}{\textbf{FaithMed 1.7B}} 
& \xmark & \xmark & None 
& 49.2 & 46.8 & 65.5 & 15.3 & 36.0 & 46.4 & 11.1 & 38.6 \\
& \xmark & \cmark & None 
& 50.9 & 50.1 & \underline{68.5} & 27.1 & 31.0 & 48.1 & 11.5 & 41.0 \\
& \xmark & \cmark & Ep. 
& 49.1 & 48.9 & 67.9 & 18.9 & \underline{39.7} & \underline{51.0} & \underline{15.0} & 41.5 \\
& \xmark & \cmark & Step 
& \textbf{61.4} & \textbf{56.9} & \textbf{75.6} & 33.2 & 38.5 & \textbf{52.0} & \textbf{17.1} & \textbf{47.8} \\
& \cmark & \xmark & None 
& 48.6 & 46.4 & 64.4 & 40.9 & 33.6 & 41.8 & 11.4 & 41.0 \\
& \cmark & \cmark & None 
& 52.4 & 51.0 & 66.9 & \underline{49.8} & \textbf{41.3} & 46.1 & 14.4 & 46.0 \\
& \cmark & \cmark & Ep. 
& 52.3 & 52.5 & 64.2 & 47.0 & 35.2 & 43.1 & 12.0 & 43.8 \\
& \cmark & \cmark & Step 
& \underline{53.5} & \underline{53.2} & 66.4 & \textbf{56.2} & 37.5 & 48.3 & 14.5 & \underline{47.1} \\
\midrule
\multirow{8}{*}{\textbf{FaithMed 4B}} 
& \xmark & \xmark & None 
& 70.6 & 63.4 & 81.5 & 29.6 & 47.3 & 55.4 & 19.6 & 52.5 \\
& \xmark & \cmark & None 
& 71.3 & 63.2 & \underline{84.4} & 39.0 & \underline{49.8} & \underline{66.1} & 22.0 & 56.5 \\
& \xmark & \cmark & Ep. 
& \underline{71.9} & \underline{63.5} & 83.2 & 41.6 & 48.7 & \textbf{66.3} & 22.5 & 56.8 \\
& \xmark & \cmark & Step 
& \textbf{74.8} & \textbf{64.1} & \textbf{85.7} & 56.5 & \textbf{55.6} & 63.5 & \textbf{28.2} & \textbf{61.2} \\
& \cmark & \xmark & None 
& 69.5 & 56.1 & 78.2 & 57.8 & 47.4 & 54.5 & 16.0 & 54.2 \\
& \cmark & \cmark & None 
& 70.4 & 58.8 & 80.2 & 60.3 & 47.0 & 66.0 & 18.9 & 57.4 \\
& \cmark & \cmark & Ep. 
& 69.4 & 59.2 & 79.7 & \underline{62.3} & 47.2 & 63.5 & 18.2 & 57.1 \\
& \cmark & \cmark & Step 
& 70.5 & 61.4 & 83.0 & \textbf{65.2} & 48.5 & 64.9 & \underline{23.0} & \underline{59.5} \\
\bottomrule
\end{tabular}%
}
\end{table*}

\section{Experiments Setup}
\label{sec:setup}
\paragraph{Datasets.}
We evaluate on a diverse suite of medical benchmarks that cover complementary types of medical tasks:
(1) knowledge-intensive question answering, including HeadQA~\citep{vilares-gomez-rodriguez-2019-head}, MedMCQA~\citep{pmlr-v174-pal22a}, and MMLU-Pro-Health~\citep{wang2024mmlupro};
(2) expert-level medical reasoning, represented by MedXpertQA~\citep{zuo2025medxpertqa};
(3) quantitative medical calculation, represented by MedCalc-Bench~\citep{khandekar2024medcalcbench}; and
(4) clinical application question answering, including MedQA~\citep{jin2021medqa} and MedBullets$^{5\mathrm{op}}$~\citep{chen-etal-2025-benchmarking}.
We use HeadQA, MedMCQA, MedCalc-Bench, and MedQA for training and in-distribution evaluation, and reserve MMLU-Pro-Health, MedXpertQA, and MedBullets$^{5\mathrm{op}}$ for out-of-distribution evaluation.
To reduce near-duplicate leakage, we apply
16-gram decontamination against the held-out evaluation sets.

\paragraph{Baselines.}
We compare \method{} against three categories of baselines:
(1) reference LLMs, including general-purpose, and medical models;
(2) evidence-utilization strategies, including MedRAG-style pre-retrieval~\citep{xiong2024benchmarkingretrievalaugmentedgenerationmedicine} and agentic search prompting; and
(3) training variants, including GRPO with episode-level outcome rewards and alternative process-reward formulations.

\paragraph{Rubrics Refinement.}
We use 300 medical QA reasoning trajectories generated by the teacher model to refine the rubric set. The refinement criteria are instantiated as follows: \textsc{Merge} is applied when any pair of rubrics has an absolute Pearson correlation above 0.70, \textsc{Rewrite} when the non-applicability rate exceeds 0.80, and \textsc{Remove} when the \textsc{Met} rate is above 0.95 or below 0.05. Starting from 26 seed rubrics, the refinement procedure converges after three iterations, yielding 15 rubrics.

\paragraph{Training Recipes.}
(1) \textit{SFT training.}
We construct supervised trajectories using DeepSeek-R1 with agentic search prompting on 2,650 questions from MedQA, MedMCQA, HeadQA, and MedCalc-Bench. For each question, we retain 6 trajectories with correct answers and valid formats; questions with fewer than 6 valid trajectories within 20 attempts are filtered out. We fine-tune Qwen3 backbones for one epoch using a learning rate of $5\times10^{-5}$, cosine decay, and a 1\% linear warmup ratio.
(2) \textit{RL training.} We perform RL training for 300 steps with a batch size of 32 and 8 rollouts per prompt. We use Gemini-2.5-Flash-Lite as the process-reward judge, set $\lambda=0.05$ for step-level process rewards, and use a maximum response length of 3072. For anchor-state grouping, we use a fuzzy string similarity threshold of 0.8. Additional implementation details are provided in the appendix \ref{app:implement-details}.

\section{Results}
\label{sec:results}
In this section, we provide a comprehensive evaluation of \method{}.
We first report benchmark performance and evidence-based medicine reasoning scores in Section~\ref{subsec:overall_results}.
We then present ablation studies over training recipes and rubric dimensions in Section~\ref{subsec:ablation_study}, followed by diagnostics of process-reward granularity and advantage grouping in Section~\ref{subsec:step_analysis}.
Finally, we present case studies that illustrate evidence-based behaviors in Section~\ref{subsec:case_study}.
\subsection{Overall Results}
\label{subsec:overall_results}
Table~\ref{tab:main-results} shows that \method{} consistently improves benchmark 
answer accuracy across both backbones. The baseline comparisons indicate that external evidence is useful, but the best access strategy depends on model capacity: MedRAG improves the 1.7B model more than agentic search (40.3\% vs. 38.6\%), 
whereas agentic search is more effective for the 4B model (52.5\% vs. 50.1\%), 
suggesting that smaller models may struggle to execute agentic workflows while benefiting more from retrieved medical knowledge. 

Beyond evidence access, outcome-only RL improves performance, but \method{} yields larger gains, 
raising benchmark answer accuracy from 41.0\% to 47.8\% on 1.7B and from 56.5\% to 61.2\% on 4B. 
This shows that learning a faithful evidence-based reasoning process improves task 
success, not only reasoning quality; notably, \method{}-4B surpasses the larger Qwen3 8B 
reference model (61.2\% vs. 60.7\%).

Beyond answer accuracy, Table~\ref{tab:rubric-results} shows that \method{}
substantially improves the evidence-based medicine reasoning process itself.
Improvements are strongest on dimensions requiring active evidence use: for both
backbones, \method{} improves Acquire by roughly 23--24\%, Appraise by roughly
21--22\%, and Apply by roughly 18\% over the base model. We also compare against
an episode-level process-reward baseline, which provides episode-level process
feedback without assigning it to specific reasoning steps. \method{} consistently
outperforms this baseline on both backbones, demonstrating the benefit of
fine-grained faithful reasoning supervision.
\method{}-4B even matches the teacher model DeepSeek-R1 on 3 out of 5 dimensions.


\subsection{Ablation Study}
\label{subsec:ablation_study}
\begin{table}[t]
\centering
\footnotesize
\setlength{\tabcolsep}{1pt}
\renewcommand{\arraystretch}{1.12}
\caption{Process reward dimension ablation results. Ablation rows report changes from the full model.}
\label{tab:dimension-ablation}
\begin{tabular*}{\columnwidth}{@{\extracolsep{\fill}}lrrrrr@{}}
\toprule
\textbf{Training} &
\textbf{Ask} &
\textbf{Acquire} &
\textbf{Appraise} &
\textbf{Apply} &
\textbf{Assess} \\
\midrule
FaithMed 1.7B & 85.8 & 50.1 & 44.9 & 38.8 & 17.5 \\
\midrule
w/o. Ask & -3.17 & -3.11 & -4.35 & -0.26 & -4.79 \\
w/o. Acquire & -0.46 & -8.28 & -5.92 & -0.11 & -4.17 \\
w/o. Appraise & -0.06 & -0.10 & -9.35 & -0.14 & -2.48 \\
w/o. Apply & +0.03 & -1.04 & -0.11 & -8.51 & -1.06 \\
w/o. Assess & -0.01 & +0.06 & -2.92 & -2.55 & -8.44 \\
\bottomrule
\end{tabular*}
\end{table}

\begin{figure*}[t]
\centering
\includegraphics[width=\textwidth]{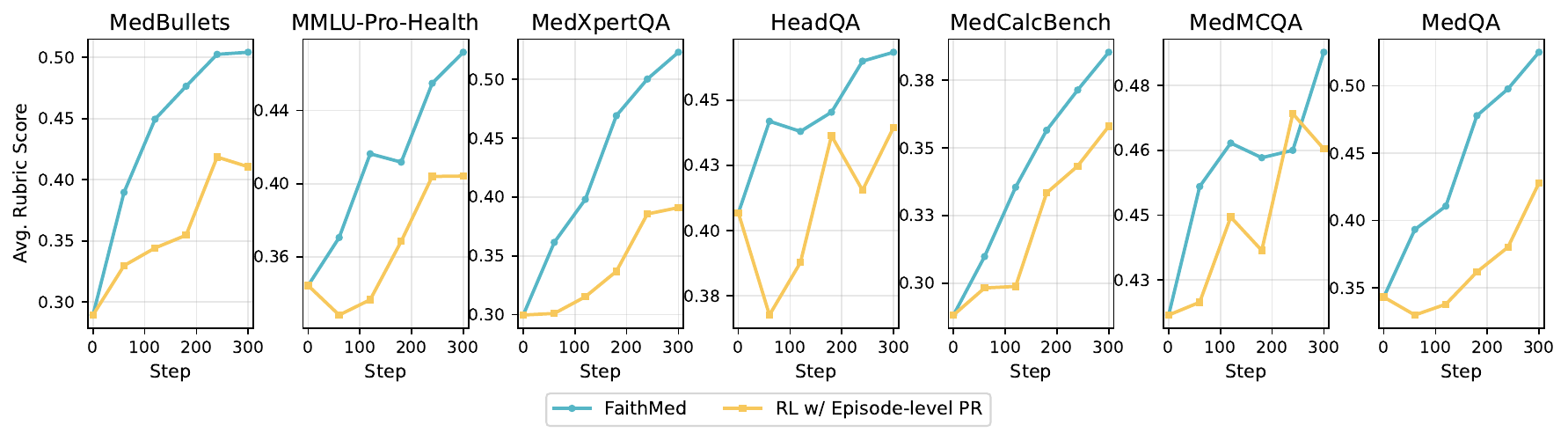}
\caption{Rubric-score curves across seven medical benchmarks. Scores are averaged over the five rubric dimensions for Qwen3-1.7B trained with SFT + RL using episode-level or step-level process reward.}
\label{fig:rubric-curve}
\end{figure*}

\begin{table*}[t]
\centering
\scriptsize
\setlength{\tabcolsep}{3pt}
\renewcommand{\arraystretch}{1.12}
\caption{Case studies of process-dimension behaviors. High-quality responses are selected from FaithMed-4B, while lower-quality responses are selected from Qwen3-4B with agentic search. Green and red highlights mark phrases most aligned or misaligned with the corresponding rubric dimension. Ellipses mark omissions made for space.}
\label{tab:case-study}
\begin{tabular}{@{}>{\raggedright\arraybackslash}p{0.11\textwidth}>{\raggedright\arraybackslash}p{0.5\textwidth}>{\raggedright\arraybackslash}p{0.34\textwidth}@{}}
\toprule
\textbf{Dimension} & \textbf{FaithMed-4B response} & \textbf{Qwen3-4B agentic-search response} \\
\midrule
Ask &
The clinical problem is to identify the metal accumulated in a patient with neuropsychiatric symptoms, dystonia, ataxia, and peripheral corneal rings. The \textcolor{checkgreen}{\textbf{decisive clue}} is the Kayser-Fleischer-type ring together with movement and psychiatric findings. This \textcolor{checkgreen}{\textbf{narrows the answer}} to Wilson disease and copper accumulation. &
The patient has tremor and abnormal gait, so this is a \textcolor{crossred}{\textbf{neurologic toxicity question}}. Metals can cause neurologic symptoms, especially lead or mercury. I will \textcolor{crossred}{\textbf{pick the metal most associated with tremor}}. \\
\midrule
Acquire &
I need the \textcolor{checkgreen}{\textbf{exact SIRS thresholds}} before counting criteria because the answer changes if respiratory rate, PaCO2, temperature, or WBC cutoffs are different. The patient's admission values are \ldots{}, so the search should target ``SIRS criteria temperature heart rate respiratory rate WBC PaCO2 thresholds.'' \texttt{<search>SIRS criteria temperature heart rate respiratory rate WBC PaCO2 thresholds</search>} &
This is about a hospitalized patient with a lung infection and abnormal labs. I should look up the \textcolor{crossred}{\textbf{underlying disease first}}: \texttt{<search> coccidioidomycosis symptoms treatment </search>}. \\
\midrule
Appraise &
The \textcolor{checkgreen}{\textbf{strongest retrieved evidence is direct}}: [T1-E1] says Kayser-Fleischer rings in Wilson disease are due to copper deposition, and [T1-E2] links excess copper deposition to neurologic and psychiatric manifestations. These passages directly answer the mechanism question, whereas passages listing broad differentials are less useful for deciding the accumulated metal. &
[T1-E5] lists several differential diagnoses for Wilson disease, including tuberculosis meningoencephalitis and CNS lymphoma, so the tissue accumulation is \textcolor{crossred}{\textbf{probably lead}}. The citation supports \textcolor{crossred}{\textbf{excluding copper}} because Kayser-Fleischer rings can occur in other diseases. \\
\midrule
Apply &
Because the child has \textcolor{checkgreen}{\textbf{4 anesthetic skin patches}}, he falls on the \textcolor{checkgreen}{\textbf{paucibacillary side}} of the lesion-count split rather than the multibacillary branch, which starts at 6 or more lesions [T1-E4]. For paucibacillary disease, the regimen is rifampicin plus dapsone [T1-E1], so options that add clofazimine are not supported \ldots{}. The dosing choice should therefore follow the PB child/adolescent regimen rather than defaulting to the adult MB-looking option. &
The patient has hypopigmented patches and sensory loss, which \textcolor{crossred}{\textbf{suggests multibacillary leprosy}}. WHO MDT includes rifampicin, dapsone, and clofazimine, so option D matches best. The \textcolor{crossred}{\textbf{adult multibacillary regimen}} is used even though the patient is 12 and has only 4 lesions. \\
\midrule
Assess &
There is a \textcolor{checkgreen}{\textbf{classification tradeoff}}: treating as multibacillary adds clofazimine and broader coverage, but the stem's 4 lesions suggests paucibacillary disease where simpler rifampicin+dapsone therapy is standard. Because \textcolor{checkgreen}{\textbf{unnecessary clofazimine increases treatment burden and adverse effects}}, the regimen should follow lesion count and age-specific guidance rather than defaulting to adult MB therapy. &
Leprosy is serious, so the \textcolor{crossred}{\textbf{strongest three-drug regimen}} is best. More antibiotics should cover more disease, and the \textcolor{crossred}{\textbf{exact number of patches does not change}} the treatment choice. \\
\bottomrule
\end{tabular}
\end{table*}

\begin{figure}[t]
\centering

\begin{subfigure}[t]{0.48\columnwidth}
    \centering
    \includegraphics[width=\linewidth]{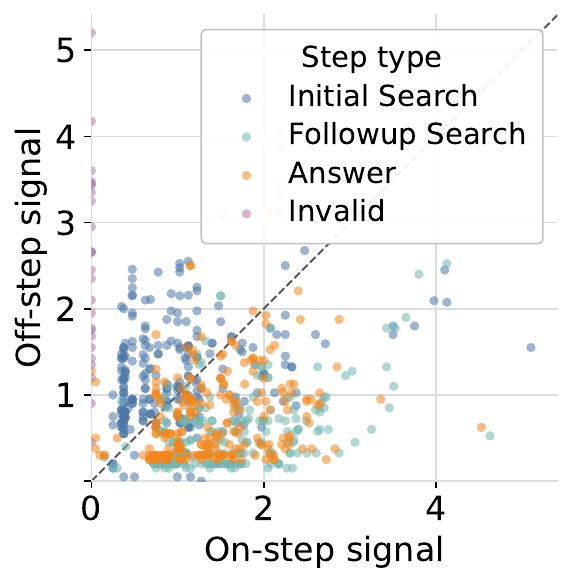}
    \caption{Episode-level signal}
    \label{fig:on_vs_off_signal}
\end{subfigure}
\hfill
\begin{subfigure}[t]{0.48\columnwidth}
    \centering
    \includegraphics[width=\linewidth]{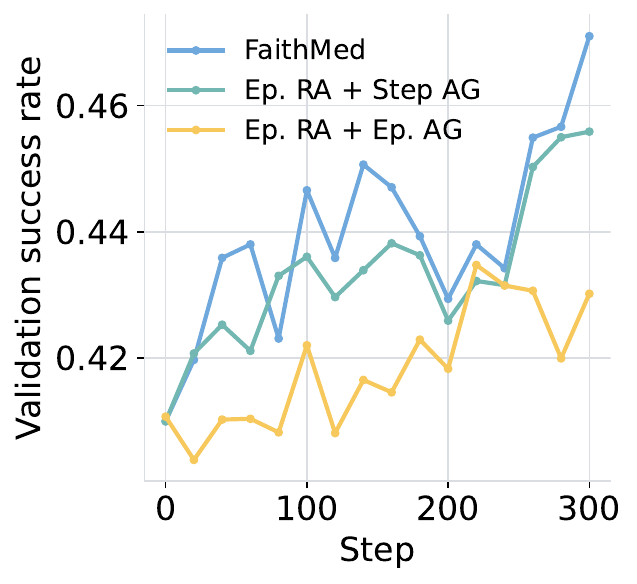}
    \caption{RA and AG ablation}
    \label{fig:val_success_rate}
\end{subfigure}
\caption{Diagnostics for reward assignment (RA) and advantage grouping (AG).}
\label{fig:training-dynamics}

\end{figure}

\paragraph{Step-level process reward is the most reliable training recipe.}
We first examine which training recipe is responsible for the gains of
\method{}. Table~\ref{tab:training-ablation} varies (1) supervised
fine-tuning and RL rewards: (2) outcome reward and (3) episode- versus
step-level process reward. Outcome-only RL improves over the untrained policy,
but the largest gains come from adding step-level process reward. Without SFT,
step-level process reward improves overall accuracy by 9.2 points for the 1.7B
backbone and 8.7 points for the 4B backbone. The same pattern holds after SFT,
where it adds 6.1 and 5.3 points over the SFT-only 1.7B and 4B models,
respectively.

Combining outcome reward with episode-level process reward is less reliable: it
improves some no-SFT settings but offers little or negative gain after SFT. 
This pattern supports a central design choice of \method{}: process supervision
should be assigned at finer granularity, so each reasoning step can be compared
with similar step groups on the corresponding evidence-based behavior.

\paragraph{Rubric dimensions target distinct reasoning stages.}
Table~\ref{tab:dimension-ablation} then ablates the reward dimensions
themselves. The largest drops occur on the ablated dimension: removing Acquire,
Appraise, Apply, and Assess reduces the corresponding score by 8.28, 9.35,
8.51, and 8.44 points, respectively. This concentration of errors shows that
the reward dimensions target distinct parts of the reasoning trace. At the same time, the
effects are not fully isolated. Removing Ask or Acquire also lowers later
stages such as Appraise and Assess, suggesting that evidence-based medical
reasoning is sequential and interdependent: poor question formulation or
evidence gathering can limit the quality of later evidence evaluation and
clinical application.

\subsection{Diagnosing Step-Level Process Reward}
\label{subsec:step_analysis}
The ablations above show that step-level process reward is more reliable than
episode-level process reward. Figure~\ref{fig:rubric-curve} confirms
this trend across benchmarks: it provides more consistent reasoning-quality
gains than the episode-level baseline. To understand why step-level process
reward works better, we diagnose this advantage through two components of
process-reward training: reward assignment (RA), which assigns
rubric feedback to steps or episodes, and advantage grouping (AG), which
normalizes advantages within anchor-state step groups or rollout groups.

\paragraph{Episode-level assignment creates off-step credit.}
Episode-level reward assignment can give a step feedback from rubrics that the
step does not directly control. Figure~\ref{fig:on_vs_off_signal} shows this
credit-assignment mismatch by decomposing the rubric signal assigned to each
step into on-step and off-step components. 
This mismatch is especially problematic for early search actions, which can be
covered by only a small number of corresponding rubrics. Their updates
can therefore be dominated by noisy off-step feedback from later evidence
appraisal or answer quality, causing the model to learn unstable signals at the
start of the reasoning process.

\paragraph{Grouping helps only with step-aligned rewards.}
To separate the role of grouping from reward assignment, we compare an
intermediate setting that uses step-level advantage grouping but keeps
episode-level reward assignment. In Figure~\ref{fig:val_success_rate}, this
setting provides only limited improvement, whereas combining step-level reward
assignment with step-level advantage grouping substantially improves
performance. This suggests that the main gain comes from aligning both the
reward signal and the comparison group with the reasoning step being optimized.

\subsection{Case Study}
\label{subsec:case_study}
\paragraph{Case studies reveal concrete reasoning differences.}
Beyond aggregate rubric scores, we examine representative trajectories to
identify what the process improvements look like in model outputs.
Table~\ref{tab:case-study} shows that the
\method{} responses explicitly identify decisive clinical clues, formulate
decision-directed searches, prioritize directly supporting evidence, and check
whether retrieved guidance applies to the patient. In contrast, the
agentic-search baseline often uses plausible but underspecified heuristics,
generic search queries, or evidence that is not logically aligned with the
clinical decision. These examples show how the rubric gains in
Table~\ref{tab:rubric-results} correspond to concrete improvements in the
reasoning trace rather than only higher final-answer accuracy.

\section{Conclusion}
\label{sec:conclusion}
We introduced \method{}, a rubric-guided training framework for faithful
evidence-based medical reasoning. Instead of rewarding only final answers,
\method{} decomposes reasoning into evidence-based medicine dimensions and uses
step-level process rewards to give credit to local
actions that ask clinically precise questions, acquire relevant evidence,
appraise retrieved passages, apply evidence to the patient context, and assess
the final decision. Across seven medical benchmarks, \method{} improves both
1.7B and 4B Qwen3 backbones and outperforms retrieval-only and agentic-search
baselines. Our ablations show that step-level process rewards provide more
consistent gains than episode-level process rewards, and rubric-level analyses
confirm that these gains reflect improved evidence-use behaviors rather than
only better answer selection. These results suggest that faithful medical
reasoning requires not only access to external knowledge, but also explicit
training of how evidence should be sought, judged, and applied.

\section{Limitations} 
\label{sec:limitation}

Our study focuses on textual medical question answering benchmarks, which
simplify the complexity of real clinical workflows involving longitudinal
records, laboratory results, imaging, physical examinations, and
institution-specific constraints. In addition, our process supervision depends
on clinician-reviewed rubrics and LLM-based judges, which are useful but still
imperfect proxies for faithful medical reasoning; models may learn to satisfy
rubric criteria without fully eliminating unsupported claims. The current
experiments are also limited to Qwen3 backbones, English medical QA settings,
and a fixed retrieval pool, so broader evaluation across model families,
languages, specialties, and evolving medical evidence sources remains future
work.

\section{Ethical Considerations} 
\label{sec:ethics}

This work is intended as research on improving the faithfulness of medical
reasoning in language models and should not be used as a substitute for
professional diagnosis, treatment planning, or patient-specific medical advice.
We evaluate on existing medical QA benchmarks and use publicly available
medical corpora for retrieval; we do not collect new patient records or personal
health information. However, models trained or evaluated in this setting can
still produce incorrect answers, misleading explanations, or unsupported
medical claims, and any real-world use would require rigorous clinical
validation, privacy safeguards for user-provided health information, and
oversight by qualified medical professionals. We also note that medical
benchmarks and retrieval corpora may encode demographic, geographic,
linguistic, or institutional biases, so improved benchmark performance does not
guarantee equitable behavior across patient populations.

\bibliography{custom}

\appendix

\section{Appendix}
\label{sec:appendix}

\subsection{Rubric Details}
\label{app:rubric-details}
We list our evidence-based medicine rubrics in Table \ref{tab:appendix-rubrics-1}, \ref{tab:appendix-rubrics-2}.  Each rubric is binary-scored as MET or UNMET with examples from teacher-generated medical QA reasoning trajectories.
\begin{table*}[t]

\centering

\scriptsize

\setlength{\tabcolsep}{3pt}

\renewcommand{\arraystretch}{1.18}

\caption{Ask and Acquire rubrics used to evaluate evidence-based medical reasoning.}

\label{tab:appendix-rubrics-1}

\begin{tabular}{@{}>{\raggedright\arraybackslash}p{0.12\textwidth}>{\raggedright\arraybackslash}p{0.27\textwidth}>{\raggedright\arraybackslash}p{0.55\textwidth}@{}}

\toprule

\textbf{Name} & \textbf{Description} & \textbf{Binary condition} \\

\midrule

\rowcolor{groupblue}

\multicolumn{3}{@{}l}{\textbf{Ask}} \\

Question clarity & Does the model contain an explicit sentence restating or paraphrasing the core clinical question, identifying the condition, mechanism, or scenario at issue? & \textbf{MET:} The \texttt{<think>} block contains a sentence such as ``The key is to choose the correct sequence of management for this genital trauma case'' or ``For this stroke patient on heparin, the task is to determine the cause of the new respiratory decline.'' \textbf{UNMET:} No concrete restatement appears, or the model only says vague phrases like ``the key point is the diagnosis'' without naming the requested target. \\

\addlinespace

Salient patient-feature extraction & Does the model identify the small set of stem features that drive the answer, rather than merely restating many details? & \textbf{MET}: The \texttt{<think>} block marks one or more features as decisive because they change the diagnosis, mechanism, calculation, management step, or option choice, e.g., ``ptosis + miosis + normal extraocular movements are the key clues for Horner syndrome''; ``catalase-positive, coagulase-positive organism is the decisive clue for Staphylococcus aureus and CGD''; ``positive pregnancy test plus LLQ pain/hypotension makes ectopic pregnancy the key concern''; or ``height 162.1 cm and weight 71.4 kg are the only values needed for BMI.'' \textbf{UNMET}: The model only paraphrases the stem, lists non-decisive details, or does not identify which features are answer-driving. \\

\addlinespace

Question specificity & Is the clinical problem narrowed to a specific question before searching or answering, rather than left as a vague topic area? & \textbf{MET}: The \texttt{<think>} block contains a precise, answerable question before \texttt{<search>} that directly maps to the decision required, e.g., ``I need to confirm whether diastolic RV collapse is specific to tamponade vs. constrictive pericarditis.'' \textbf{UNMET}: The model only states the broad task type without naming the decision boundary, such as ``answer a thyroid nodule question'' or ``calculate creatinine clearance.'' \\

\midrule

\rowcolor{groupblue}

\multicolumn{3}{@{}l}{\textbf{Acquire}} \\

Search Motivation and Alignment & Does the model explicitly identify a clinically relevant knowledge gap, unresolved uncertainty, or claim that genuinely needs verification, and does its search directly target that gap? & \textbf{MET}: Immediately before \texttt{<search>}, the model names one specific unresolved decision-critical gap, such as a diagnostic discriminator, legal/ethical rule, dosing formula, threshold, or comparison between answer options, and the query explicitly contains the terms needed to resolve that exact gap. A high-level topic query is not enough: the query should encode the actual decision boundary named in \texttt{<think>}, not just the domain. \textbf{UNMET}: The model has already mostly committed to an answer and the search is only decorative confirmation, or it searches for information it already states with confidence and does not explain why verification is still necessary. \\

\addlinespace

Gap-directed search progression & This rubric requires at least one prior retrieval to assess progress. Compared to information retrieved in prior turns, does the current search make progress toward resolving the outstanding evidence need, rather than repeating resolved topics or drifting to unrelated areas? & \textbf{MET}: The \texttt{<think>} block names what the previous results did not resolve and searches more specifically, e.g., ``The first search returned general thyroid malignancy features, but I need the official ACR scoring categories'' or ``The results mention treatment but not renal-failure dosing, so I should search dosing in renal impairment.'' \textbf{UNMET}: No explicit gap-directed search progression pattern appears. \\

\addlinespace

Query constraint preservation & Does the search query preserve the answer-critical constraints from the question stem, such as patient subgroup, timing, comparator, option term, disease subtype, formula variable, or requested source standard? & \textbf{MET}: The query includes the key constraint that would change the answer if omitted, e.g., ``complications of meningococcal meningitis in children'' or ``tetanus prophylaxis in patients taking glucocorticoids.'' \textbf{UNMET}: No explicit query-constraint-preservation pattern appears. \\

\bottomrule

\end{tabular}

\end{table*}
\begin{table*}[!t]
\centering
\scriptsize
\setlength{\tabcolsep}{3pt}
\renewcommand{\arraystretch}{1.18}
\caption{Appraise, Apply, and Assess rubrics used to evaluate evidence-based medical reasoning. Each rubric is binary-scored as MET or UNMET with examples from teacher-generated medical QA reasoning trajectories.}
\label{tab:appendix-rubrics-2}

\begin{tabular}{@{}>{\raggedright\arraybackslash}p{0.12\textwidth}>{\raggedright\arraybackslash}p{0.27\textwidth}>{\raggedright\arraybackslash}p{0.55\textwidth}@{}}
\toprule
\textbf{Name} & \textbf{Description} & \textbf{Binary condition} \\
\midrule

\rowcolor{groupblue}
\multicolumn{3}{@{}l}{\textbf{Appraise}} \\

Evidence-faithful synthesis & Do the clinical claims attributed to specific [Tn-Rn] passages accurately reflect the content of those passages, with the correct logical direction and without inferential jumps not shown in \texttt{<think>}? & \textbf{MET}: Every claim that cites a [Tn-Rn] passage (a) can be verified by reading that passage, (b) preserves the direction of the cited relationship, such as causality, magnitude, or association, and (c) requires no unstated inferential step between the passage text and the attributed claim. \textbf{UNMET}: At least one [Tn-Rn] citation violates any of these conditions: the passage does not support the attributed claim, the direction of the cited relationship is reversed or distorted, or the attributed claim requires an inferential step not shown in \texttt{<think>}. \\

\addlinespace
Evidence conflict recognition & When the model notices inconsistent or qualifying retrieved information, does it explicitly name the discrepancy? & \textbf{MET}: The \texttt{<think>} block contains a clear conflict-recognition pattern comparing retrieved passages, e.g., ``[T1-R2] says two tetanus doses during pregnancy, while [T1-R3] says complete a three-dose primary series; the guidance differs.'' \textbf{UNMET}: The conflict is invisible in the reasoning. \\

\addlinespace
Evidence sufficiency judgment & When the retrieved [Tn-Rn] passages do not directly address the clinical concept targeted by the preceding \texttt{<search>}, does \texttt{<think>} explicitly state this insufficiency before proceeding to draw conclusions? & \textbf{MET}: The \texttt{<think>} block contains a sentence such as ``the retrieved passages do not directly confirm polonium's reactivity with HCl,'' ``no passage specifically addressed dosing in renal failure'' before reasoning forward from indirect evidence, or ``The retrieved passages do not directly address renal dosing, so the answer remains tentative.'' \textbf{UNMET}: The \texttt{<think>} block does not contain any explicit statement acknowledging that the retrieved passages fail to directly address the search target. \\

\midrule

\rowcolor{groupblue}
\multicolumn{3}{@{}l}{\textbf{Apply}} \\

Setting applicability & Does the model explicitly assess whether at least one key [Tn-Rn] passage applies to the specific patient or scenario described in the question stem, by comparing the passage's population or context to the patient's features? & \textbf{MET}: The \texttt{<think>} or \texttt{<answer>} block contains an explicit applicability statement, e.g., ``T2-R1 reports echocardiographic findings in ICU patients with effusion, which matches the ED presentation in this case,'' ``T1-R3 is about neonates born to infected mothers, directly applicable here,'' or ``The CGD evidence applies because the patient has recurrent catalase-positive infections.'' \textbf{UNMET}: No explicit applicability pattern appears. \\

\addlinespace
Evidence-based exclusion of alternatives & Does the model ground its option exclusion reasoning in [Tn-Rn] citations in \texttt{<think>}, rather than dismissing options by parametric assertion alone? & \textbf{MET}: Option elimination in \texttt{<think>} is explicitly tied to a [Tn-Rn] citation, e.g., ``[T1-R2] confirms pericardial thickening is the hallmark of constrictive pericarditis, ruling out option B in this patient whose echo shows no thickening'' or ``[T1-R6] lists margin, echogenicity, and shape as ACR features, leaving vascularity as the exception.'' \textbf{UNMET}: Option elimination in \texttt{<think>} relies on bare assertion with no [Tn-Rn] citation. \\

\addlinespace
Evidence Substantiation and Prioritization & Does the model explicitly prioritize the most direct or authoritative retrieved evidence when supporting the answer? & \textbf{MET}: The reasoning uses retrieved evidence as the main basis for the answer and names why the chosen evidence is more direct, authoritative, or answer-relevant than alternatives, e.g., ``The official ACR TI-RADS feature list is more directly relevant than a general malignancy-feature review, so I will use [T1-R6]''; ``CDC guideline language should outweigh a broad background article for vaccination scheduling''; or ``[T1-R1] and [T1-R5] directly address radiolucency of uric acid stones, while composition-only passages are less useful for this question.'' \textbf{UNMET}: No explicit evidence-prioritization or authority-evaluation pattern appears. \\

\midrule

\rowcolor{groupblue}
\multicolumn{3}{@{}l}{\textbf{Assess}} \\

Uncertainty calibration & Does the model explicitly acknowledge limitations in the retrieved information or express appropriate uncertainty before drawing a conclusion? & \textbf{MET}: At least one explicit hedge appears in \texttt{<think>} or \texttt{<answer>} whenever the evidentiary basis is weak, e.g., ``the retrieved passages do not confirm X; based on clinical reasoning I tentatively conclude Y,'' or ``limited data support this -- evidence suggests...'' in \texttt{<answer>} when the trajectory contained only peripheral [Tn-Rn] passages. \textbf{UNMET}: No explicit uncertainty/calibration pattern appears. \\

\addlinespace
Premature closure self-check & Does the model avoid locking onto an early answer by explicitly checking and revising a plausible but wrong hypothesis before committing? & \textbf{MET}: The \texttt{<think>} block contains a visible self-correction pattern: it names an initial candidate or tempting interpretation, identifies the specific stem clue or retrieved evidence that weakens it, and shifts to the better-supported answer, e.g., ``I initially thought this was tension pneumothorax because of tracheal deviation, but the percussion is dull rather than hyperresonant, so that hypothesis fails and hemothorax is the answer'' or ``I was leaning toward opioid intoxication, but opioid toxicity should cause miosis and respiratory depression, whereas this stem has dry mucous membranes, agitation, hypertension, and mydriasis, so I should switch to anticholinergic toxicity.'' \textbf{UNMET}: No explicit hypothesis-checking or self-correction pattern appears before the final answer. \\

\addlinespace
Tradeoff communication & Does \texttt{<think>} or \texttt{<answer>} explicitly name at least one relevant tradeoff, such as benefit--harm, efficacy--tolerability, or convenience--monitoring burden? & \textbf{MET}: The \texttt{<think>} or \texttt{<answer>} block discusses at least one tradeoff relevant to the recommendation, e.g., ``[T1-R4] reports ATA has higher sensitivity for thyroid nodule biopsy decisions, while [T1-R5] reports ACR TI-RADS has higher specificity and fewer unnecessary FNAs; this is a sensitivity-specificity tradeoff rather than a single uniform recommendation'' or ``[T1-R6] shows TXA is more effective than packing in antiplatelet users, but IV administration adds procedural burden compared to topical oxymetazoline.'' \textbf{UNMET}: No explicit tradeoff-communication pattern appears. \\

\bottomrule
\end{tabular}
\end{table*}

\subsection{Implementation Details}
\label{app:implement-details}
\textbf{Retriever Implementation.}
We build the retrieval pool from five medical corpora. Four corpora---PubMed,
StatPearls, medical textbooks, and Wikipedia---are taken from the MedRAG
benchmark~\citep{xiong2024benchmarkingretrievalaugmentedgenerationmedicine}. We
additionally include the clinical guideline corpus released with
MediTron-70B~\citep{chen2023meditron70bscalingmedicalpretraining}. We segment
all source documents into short snippets and encode them with
MiniCPM-Embedding-Light using the Tevatron dense-retrieval
pipeline~\citep{gao2022tevatron}. The resulting snippet embeddings are sharded
and served with DiskANN indices for efficient vector search~\citep{subramanya2019diskann}.\\
\newline
\textbf{RL Implementation.} 
We denote the score for rubric
$m\in\mathcal{R}_k$ as $J_m(y_k,e_{k-1},\mathcal{R}_k)$.
In our retrieval setting, the anchor state
$\tilde{s}_{i,k}$ for step $k$ in trajectory $i$ is the observation-induced
state used for grouping: $\tilde{s}_{i,1}=x$ for the initial step, and
$\tilde{s}_{i,k}=e_{i,k-1}$ for later steps. For the $N$ trajectories sampled
for the same question $x$, the step-level group anchored at $\tilde{s}$ is
\begin{equation}
\mathcal{G}(\tilde{s})=
\{(i,k): \mathrm{sim}(\tilde{s}_{i,k},\tilde{s})\geq \eta\},
\end{equation}
where $\mathrm{sim}(\cdot,\cdot)$ measures anchor-state similarity and $\eta$
is the grouping threshold. Thus, all initial steps for the same question form one
group, and later steps are compared with other steps that condition on similar retrieved evidence.

Before summing rubric scores, we center each rubric within the local group to
remove rubric-specific baselines. Let $\mu_{i,k,m}$ denote the mean score of
rubric $m$ among group members for which $m$ is applicable. The step-level
process reward is then the sum of centered rubric scores:
\begin{equation}
\label{eq:app-process-reward}
r^P_{i,k}=\sum_{m\in\mathcal{R}_{i,k}}
\left(J_m(y_{i,k},e_{i,k-1},\mathcal{R}_{i,k})-\mu_{i,k,m}\right).
\end{equation}
This rubric-level centering reduces sensitivity to rubric-specific positive
rates before scores are accumulated into a step reward. We then compute the
process advantage with a group-wise scale factor:
\begin{equation}
\label{eq:app-process-advantage}
A^{P}_{i,k}=
\frac{
r^P_{i,k}-\mathrm{mean}(\{r^P_{j,l}:(j,l)\in\mathcal{G}(\tilde{s}_{i,k})\})
}{
\mathrm{std}(\{r^P_{j,l}:(j,l)\in\mathcal{G}(\tilde{s}_{i,k})\})+\epsilon
}.
\end{equation}
This makes the process feedback comparative: a search, or
answer step is rewarded when it better satisfies the relevant rubrics than
alternative steps taken under a similar observation context.\\
\newline
\noindent\textbf{Supervise Finetuning.} We construct supervised trajectories using DeepSeek-R1 from 1k MedQA, 0.5k
HeadQA, 1k MedMCQA, and 0.5k MedCalc-Bench training questions. We apply rejection sampling by retaining only trajectories with correct answers and valid output formats. For each question, we collect up to 6 valid trajectories; questions with fewer than 6 valid trajectories after 20 attempts are filtered as hard questions. In total, the SFT data generation process yields 2,650 questions, 15,900 trajectories, and 28,368 SFT training samples, where each reasoning step is treated as one training sample. We additionally collect 7,538 rejected trajectories and trajectories from hard questions.\\
\newline
\textbf{Reinforcement Learning.} We implement RL training using verl-agent, an agent-training extension of verl~\citep{sheng2025hybridflow,feng2025gigpo}. Training is performed for 300 steps with a batch size of 32 and 8 rollouts per prompt. For advantage grouping, we follow GiGPO~\citep{feng2025gigpo}. For process-reward scoring, we use Gemini-2.5-Flash-Lite as the LLM judge. Rubric evaluation is performed with a single batched judge call per reasoning step instead of scoring rubrics independently, improving training efficiency. Training takes approximately 32 hours on 4 H100 GPUs.\\
\newline
\textbf{Process Reward Aggregation.} We first mean-center each rubric score within its applicable subset, since some rubrics require specific preconditions such as search actions, citations, or evidence from multiple sources. This prevents the model from favoring trajectories that only satisfy easier or more frequently applicable rubrics. The process reward for each reasoning step is then computed as the mean of centered scores across all applicable rubrics.

\newcounter{prompt}
\renewcommand{\theprompt}{\Alph{section}.\arabic{prompt}}

\subsection{Agentic Search Prompts}
\label{app:agentic-search-prompts}
We use the following prompt templates to elicit multi-turn agentic search. 

\paragraph{Medical QA benchmark initial turn template.}
The template used for all medical QA benchmarks with no previous turn
(search) histories at turn 1.
\refstepcounter{prompt}\label{prompt:medqa-no-history}
\begin{promptbox}{Prompt~\theprompt: Medical QA initial agentic search template}
You are a medical expert AI assistant. You have access to an
external search tool to retrieve information. You can answer a
question with many turns of search and reasoning.

Based on your search history, you need to take the next action to
complete the task. You will be provided with:
1. Your history search attempts: query in format
   <search> query </search> and the search results returned by
   the external search tool.
2. The question to answer.

Valid actions (ONLY TAKE ONE of the following):
(1) <search> your query </search>: search to verify a specific
    medical fact, diagnosis, treatment, drug mechanism, or clinical
    guideline relevant to the question. Your turn ends here; do not
    fabricate search results -- the external search tool will return
    them in the next turn.
(2) <answer> your answer, with the final choice in \boxed{}
    </answer>: output the final answer once you have verified
    the key medical facts.

Strategy: Think step by step about what specific medical knowledge
is needed to answer this question, then search to verify it.

Note:
(1) DO NOT simulate or imagine search results in your reasoning. If
    you decide to search, you must output a <search> action.
(2) Output EXACTLY ONE action block (search or answer).

Question: {task_description}
\end{promptbox}

\paragraph{MedCalc-Bench template.}
\refstepcounter{prompt}\label{prompt:medcalc-delta}
For MedCalc-Bench questions, we reuse the same agentic-search structure but
replace the task-specific search and answer instructions with the calculation
variants in Prompt~\ref{prompt:medcalc-delta}. The value format
instruction follows the Medmarks-Verifiable evaluation protocol
\citep{warner2026medmarks}.

\begin{promptbox}{Prompt~\theprompt: MedCalc-Bench template}
<search> your query </search>: search to verify a specific medical
formula, reference value, drug dose, or clinical parameter needed
to solve this calculation. Your turn ends here; do not fabricate
search results -- the external search tool will return them in the
next turn.

<answer> your answer, with the final value in \boxed{} </answer>:
output the final answer once you have verified the formula or key
values.

Think step by step about which formula or reference value is
required, then search to confirm it.

Use one of the following final-value formats:
- Decimal Answer Format: 17.29
- Score-Based Answer Format: 5
- Estimated Date Answer Format: 5/21/2021
- Estimated Age Answer Format: (4 weeks, 3 days)
\end{promptbox}

\paragraph{Evidence grounding prompt.}
\refstepcounter{prompt}\label{prompt:history-citation-delta}
At the first turn, the model receives an empty history and no citation
requirement, since no retrieved evidence has been observed. After the search tool
returns at least one real \texttt{<information>} block, we append the history
field and the citation rule in Prompt~\ref{prompt:history-citation-delta}. The
citation rule is used only to ground the model's internal reasoning over prior
retrieval results; the model is still required to output exactly one action
block.

\begin{promptbox}{Prompt~\theprompt: Evidence grounding prompt}
1. Your history search attempts: query in format
   <search> query </search> and the returned search results
   in <information> and </information>.

Citation rule (ONLY for internal thinking):
(1) Whenever you draw on a fact from prior search results
    (from <information> blocks), cite it inline using
    [T<turn>-R<result>] (example: [T2-R3]) in the thinking
    process.
(2) If no prior <information> block exists, or the retrieved
    information is irrelevant or noisy, do not output any
    citations.
(3) Do not invent citation ids.

For MedCalc-Bench calculation questions, cite retrieved facts,
formula inputs, or numeric assumptions in the same format.

Text between <information></information> is the search results
from the search engine after you perform a search action.
DO NOT include any information in <information></information>
in your output.

History Turns: {memory_context}
\end{promptbox}

\paragraph{Format reminder.}
\refstepcounter{prompt}\label{prompt:format-reminder}
When a model emits an invalid action, we provide the lightweight repair prompt
in Prompt~\ref{prompt:format-reminder}. This prompt does not introduce new task
information; it only restates the required action schema so that the next turn
can be parsed by the agent environment.

\begin{promptbox}{Prompt~\theprompt: Format reminder}
You generated an invalid action in your previous turn.
Output EXACTLY ONE action block:
(1) If taking a search action, please use format:
    <search> your query </search>. Do not fabricate search
    results; the external search tool will return them in the
    next turn.
(2) If taking an answer action, please use format:
    <answer> your answer, with the final choice in \boxed{}
    </answer>.
\end{promptbox}

\subsection{LLM-judge Evaluation}
\label{subsec:judge_robustness}
\begin{table}[!htbp]
\centering
\footnotesize
\setlength{\tabcolsep}{0pt}
\caption{Meta-evaluation of LLM judges across five evidence-based medicine dimensions. Gemini 2.5 and Gemini 3 denote Gemini-2.5-Flash-Lite and Gemini-3-Flash, respectively. Agree. denotes inter-annotator agreement between the two judges.}
\label{tab:judge-robustness}
\begin{tabular*}{\columnwidth}{@{\extracolsep{\fill}}lcccc@{}}
\toprule
Dimension &
\begin{tabular}[c]{@{}c@{}}Gemini 2.5\\Score\end{tabular} &
\begin{tabular}[c]{@{}c@{}}Gemini 3\\Score\end{tabular} &
Agree. &
\begin{tabular}[c]{@{}c@{}}Cohen's\\$\kappa$\end{tabular} \\
\midrule
Ask      & 0.90 & 0.89 & 0.93 & 0.62 \\
Acquire  & 0.54 & 0.57 & 0.71 & 0.41 \\
Appraise & 0.48 & 0.44 & 0.79 & 0.58 \\
Apply    & 0.41 & 0.42 & 0.77 & 0.54 \\
Assess   & 0.19 & 0.21 & 0.82 & 0.45 \\
\midrule
Macro Avg. & 0.50 & 0.51 & 0.80 & 0.52 \\
\bottomrule
\end{tabular*}
\end{table}

Table~\ref{tab:judge-robustness} compares rubric scores from Gemini 2.5 and
Gemini 3. The two judges produce nearly identical macro-average rubric scores
(50\% vs.\ 51\%) with 80\% average agreement, suggesting that the rubric labels
are robust across judges. Agreement is highest for Ask (93\%) and lower for
Acquire and Apply, which require more subjective assessment of search and
evidence use. The average Cohen's $\kappa$ is 0.52, indicating moderate
agreement beyond label imbalance and supporting the reliability of our
process-quality analysis.

\end{document}